\documentclass[10pt,twocolumn,letterpaper]{article}
\usepackage{ifthen}
\usepackage{wacv}

\usepackage{graphicx}
\usepackage{algorithm}
\usepackage{algorithmic}
\usepackage{booktabs}
\usepackage{mathrsfs}
\usepackage{color}
\usepackage{textcomp}
\usepackage{bbding}
\usepackage{pifont}
\usepackage{wasysym}
\usepackage{xcolor,colortbl}
\usepackage{overpic}
\usepackage{arydshln}
\usepackage{microtype}
\usepackage{multirow}
\usepackage{makecell}
\usepackage{hhline}
\usepackage{cuted}
\usepackage{ragged2e}
\usepackage{amsmath}

\definecolor{wacvblue}{rgb}{0.21,0.49,0.74}
\usepackage[breaklinks,colorlinks=false]{hyperref}

\def\wacvPaperID{2709} 
\def\confName{WACV}
\def\confYear{2026}

\title{SpecGen: Neural Spectral BRDF Generation via Spectral-Spatial \\ Tri-plane Aggregation}

\author{
Zhenyu Jin\thanks{Equal contribution.}  , 
Wenjie Li\footnotemark[1]  ,
Zhanyu Ma,
Heng Guo\thanks{Corresponding author.} \\
Beijing University of Posts and Telecommunications \\
{\tt\small \{sosjzy, cswjli, mazhanyu, guoheng\}@bupt.edu.cn}
}

\begin{document}

\newboolean{putfigfirst}


	
        


\maketitle

\begin{abstract}   

Synthesizing spectral images across different wavelengths is essential for photorealistic rendering. Unlike conventional spectral uplifting methods that convert RGB images into spectral ones, 
we introduce \textbf{SpecGen}, a method that generates spectral bidirectional reflectance distribution functions (BRDFs) from a single RGB image of a sphere. 
This enables spectral image rendering under arbitrary illuminations and shapes covered by the corresponding material. 
A key challenge in spectral BRDF generation is the scarcity of measured spectral BRDF data. 
To address this, we propose the Spectral-Spatial Tri-plane Aggregation (SSTA) network, which models reflectance responses across wavelengths and incident-outgoing directions, allowing the training strategy to leverage abundant RGB BRDF data to enhance spectral BRDF generation. 
Experiments show that our method accurately reconstructs spectral BRDFs from limited spectral data and surpasses existing methods in hyperspectral image reconstruction, achieving an improvement of 8\,dB in PSNR. 
Codes link: 
\href{https://github.com/sosjzy/SpecGen}{\texttt{https://github.com/sosjzy/SpecGen}}.

\end{abstract}

\section{Introduction}
Real light is composed of a spectrum of wavelengths, and photorealistic image synthesis requires capturing the wavelength-dependent reflectance properties of materials under varying illumination and viewing conditions. Spectral bidirectional reflectance distribution functions (BRDFs) constitute a crucial ingredient in this process, enabling the synthesis of hyperspectral images for remote sensing~\cite{ghassemian2016review,cheng2022multi}, material analysis~\cite{choudhary2022recent}, and virtual reality~\cite{kang2020deephandsvr}.
However, acquiring high-quality spectral BRDF data is challenging, as the measurement process requires scanning a four-dimensional domain at high resolution—a tedious and time-consuming task~\cite{dupuy2018adaptive}, leading existing spectral BRDF datasets to be scarce and limited in scale. To bypass the need for extensive spectral measurements, it is desired to develop a spectral BRDF generation method that can produce photorealistic spectral BRDFs from a single image input.

\begin{figure*}
\vspace{-0.3in}
\begin{overpic}[width=0.99\linewidth]{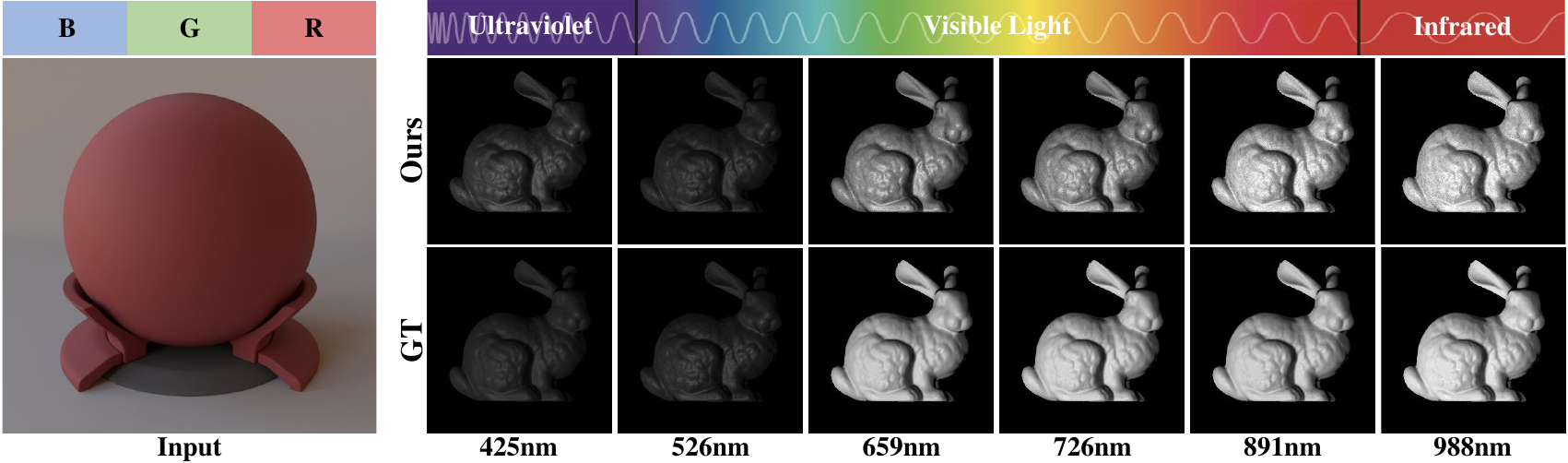}
\end{overpic}
\captionof{figure}{From a single RGB image of a sphere~(\textbf{left}), our method recovers the spectral BRDF of the corresponding material, allowing the spectral image rendering of arbitrary illuminations and shapes covered by the material under different wavelengths (\textbf{right}).}
\label{fig: teaser}
\end{figure*}%

Most existing BRDF generation methods target RGB BRDFs. For instance, Gokbudak~\etal~\cite{gokbudak2025hypernetworks} use a hypernetwork for sparse RGB BRDF reconstruction, while ControlMat~\cite{vecchio2024controlmat} and MatFuse~\cite{vecchio2024matfuse} employ diffusion models to predict spatially varying BRDF maps from a single flash RGB photo. These approaches, though effective for RGB, cannot be directly extended to spectral BRDFs due to representation differences and scarce spectral data. Separately, spectral uplifting methods~\cite{li2017hyperspectral,Wang_2019_CVPR,liu2024physics,chen2023spectral} estimate spectral images from RGB, but do not recover spectral BRDFs, hindering appearance transfer across shapes and lighting. Thus, despite advances in RGB BRDF and spectral image generation, spectral BRDF generation remains unsolved.

The key challenge of spectral BRDF generation lies in the scarcity of spectral BRDF data. Currently, only 51 isotropic spectral BRDFs are publicly available~\cite{dupuy2018adaptive}, which makes it difficult to train high-quality, generalizable models. In contrast, measured RGB BRDF datasets, such as MERL~\cite{matusik2003data}, are relatively easy to collect and are available at a much larger scale. Our key insight is to design a compatible spectral BRDF representation, along with a training strategy that leverages large-scale RGB BRDF data to facilitate spectral BRDF generation.

In this paper, we propose SpecGen, which, to the best of our knowledge, is the first network to generate spectral BRDFs from a single RGB image. The core of SpecGen is the Spectral-Spatial Tri-plane Aggregation (SSTA) module, which facilitates the mapping between RGB and spectral BRDFs. Specifically, spectral BRDFs can be represented as a 4D function of $({\theta}_h, {\theta}_d, {\varphi}_d, \lambda)$, where spatial attributes $({\theta}_h, {\theta}_d, {\varphi}_d)$—related to incident and outgoing directions—are shared between spectral and RGB BRDFs, while the spectral attribute $\lambda$ (wavelength) is unique to spectral BRDFs. Inspired by the 4D function decomposition in dynamic neural radiance fields $(x, y, z, t)$, we decompose spatial and spectral attributes into two tri-planes, encoding their corresponding latent features separately. Building upon this tri-plane representation, we design a training strategy that leverages large-scale RGB BRDF datasets to supervise the spatial tri-planes, thereby alleviating the data scarcity issue in the spectral domain. Furthermore, to effectively integrate RGB and spectral features, we propose an Adaptive Feature Fusion (AFF) module, which replaces traditional feature fusion such as Hadamard product~\cite{fridovich2023k}. The AFF module enhances the model's representation by adaptively combining multi-plane features. As shown in Fig.~\ref{fig: teaser}, these dedicated designs enable our method to accurately generate spectral BRDFs and to facilitate downstream spectral rendering tasks. Our main contributions are as follows:
\begin{itemize}
\item SpecGen is, to the best of our knowledge, the first spectral BRDF generation framework capable of generating photorealistic spectral BRDFs from a single RGB image.
\item We introduce the SSTA module, which alleviates the scarcity of spectral BRDF data by leveraging large-scale RGB BRDF datasets through a novel tri-plane representation of spectral and spatial BRDF attributes.
\item We present the AFF module, which dynamically selects and fuses RGB and spectral features to enhance the model’s representational capacity.
\end{itemize}
Extensive experiments demonstrate that our method can reconstruct and generate high-quality spectral BRDFs, benefiting downstream tasks such as hyperspectral image rendering.

\section{Related work}
As shown in Table~\ref{tab:comparison}, we summarize existing methods for BRDF generation and spectral image reconstruction from RGB images. Since our method is based on neural implicit representations, we briefly review relevant works on spectral BRDF generation and neural implicit representations.
\begin{table*}
          \caption{Summary of existing representative methods on BRDF generation.}
          \vspace{-1mm}
          \label{tab:comparison}
          \resizebox{0.98\textwidth}{!}{
          \begin{tabular}{lcccc}
          \toprule
          \textbf{Method} & \textbf{Input} & \textbf{Output} & \textbf{Main Technology} & \textbf{Generate spectral BRDF}\\
          \midrule
          ControlMat~\cite{vecchio2024controlmat}, MatFuse~\cite{vecchio2024matfuse} &RGB Image& RGB SVBRDF& Diffusion Model & \XSolid   \\
          PhotoMat~\cite{zhou2023photomat}, TileGen~\cite{zhou2022tilegen} &Flash RGB Image& RGB SVBRDF & GAN & \XSolid \\
          Metappearance~\cite{fischer2022metappearance}& Image \& Light & RGB BRDF & Meta Learning & \XSolid \\
          Gokbudak \etal~\cite{gokbudak2025hypernetworks} &BRDF Sampling Data& RGB BRDF &Hyper Network & \XSolid \\
          Narumoto \etal~\cite{narumoto2024synthesizing} &BRDF Sampling Data& Time-varying RGB BRDF & Coordinate-based MLP & \XSolid \\
          \textbf{Ours} &Image& \makecell{(RGB+Spectral) BRDF} & Triplane Representation & \Checkmark\\
          \bottomrule
          \end{tabular}}
\end{table*}
\subsection{BRDF Generation}
BRDF reconstruction and generation play a key role in advancing computer graphics and material analysis. Early methods, such as those based on CNNs~\cite{deschaintre2020guided, li2017modeling}, extract spatially varying BRDFs (SVBRDFs) from images, but struggle with generalization and require separate training for each material. Later works utilize U-Net architectures~\cite{deschaintre2019flexible, deschaintre2018single}, which improve generalization performance but often fail to preserve fine details, particularly in highlight regions. To address this limitation, GAN-based methods~\cite{vecchio2021surfacenet,guo2020materialgan,zhou2022tilegen} have gained popularity. For instance, SurfaceNet~\cite{vecchio2021surfacenet} and MaterialGAN~\cite{guo2020materialgan} utilize GAN structures to generate realistic SVBRDFs from a single image. TileGen~\cite{zhou2022tilegen} extends MaterialGAN by training category-specific models, while PhotoMat~\cite{zhou2023photomat} builds upon prior works by incorporating real-world datasets to further enhance visual quality.

With the rise of diffusion models~\cite{ho2020denoising}, recent works such as ControlMat~\cite{vecchio2024controlmat} and MatFuse~\cite{vecchio2024matfuse} have applied them to BRDF generation, effectively reducing the low-quality artifacts common in GAN-based approaches but at the expense of markedly slower generation speeds. To improve efficiency, meta-learning methods~\cite{fischer2022metappearance, zhou2022look} enable rapid inference within limited computational budgets, though this often comes with a noticeable drop in accuracy. Despite these advances, all above approaches focus exclusively on RGB BRDFs; to our knowledge, ours is the first to enable both RGB and spectral BRDF representation and generation.

Most existing BRDF generation approaches rely on parametric reflectance models~\cite{burley2012physically}, parameterizing reflectance via attributes such as albedo, roughness, and metallicity. Another line of work explores measured BRDFs from datasets like MERL~\cite{matusik2003data} and RGL~\cite{dupuy2018adaptive}. For example, Matusik~\etal~\cite{matusik2003data} used PCA to build a data-driven BRDF model; Gokbudak~\etal~\cite{gokbudak2025hypernetworks} proposed a continuous neural representation for sparse reconstruction of unseen materials; and Narumoto~\etal~\cite{narumoto2024synthesizing} developed a neural embedding with a temporal model to capture BRDF changes in latent space. However, these also concentrate on RGB BRDFs, leaving spectral BRDF generation largely unexplored.

\subsection{Implicit Neural Representations}
Unlike discrete representations such as meshes, voxels, or point grids—which store sampled values at fixed locations, leading to high memory usage, aliasing, and the need for explicit interpolation \cite{hanocka2019meshcnn}—implicit neural representations (INRs) use a learnable function (typically an MLP) to map each continuous coordinate in the signal domain to its value. This design makes INRs resolution-independent, differentiable, and far more memory-efficient than dense grids. These advantages have been demonstrated in various domains: SIREN \cite{sitzmann2020implicit} employs sinusoidal activations to capture high-frequency signals and derivatives; NeRF \cite{mildenhall2021nerf} models static scenes as continuous 5-D radiance fields for photorealistic novel-view synthesis; and Neural Descriptor Fields (NDFs) \cite{simeonov2022neural} encode object points with relative poses for smooth interpolation of robot grasp trajectories.

Building on this foundation, K-Planes~\cite{fridovich2023k} replaces the monolithic MLP with a set of locally parameterized feature planes, thereby reducing the memory requirement from $O(N^4)$ to $O(6N^2)$ for four-dimensional signals and enabling asymmetric supervision across coordinate subsets. Inspired by K-Planes, our SpecGen introduces a novel spectral and spatial triplane representation to effectively decompose and model spectral BRDFs, thereby addressing the challenge of data scarcity in spectral BRDF generation.

\subsection{Hyperspectral Image Reconstruction}
Hyperspectral image (HSI) reconstruction seeks to recover a full spectral data cube—from inputs with limited spectral resolution such as RGB images. Li~\etal~\cite{li2017hyperspectral} showed that deep CNNs can enhance textures and synthesize plausible spectra; Wang~\etal~\cite{Wang_2019_CVPR} improved robustness via optimization with data-driven priors; and Cai~\etal~\cite{cai2022mask} used a mask-guided spectral transformer with self-attention to capture long-range spatial–spectral relationships, achieving the best results.

However, existing HSI methods~\cite{li2017hyperspectral,Wang_2019_CVPR,cai2022mask} operate solely in the image domain, uplifting per-pixel spectra without disentangling material reflectance from lighting and geometry, preventing transfer to new shapes, viewpoints, or illumination. This limits applications such as relighting, compositing, and physically based simulation. In contrast, our spectral-BRDF paradigm explicitly models bidirectional, wavelength-dependent material reflectance, enabling accurate spectral rendering on arbitrary geometries and arbitrary lighting.

\section{Method}
\begin{figure*}[ht]
	\begin{overpic}[width=0.99\linewidth]{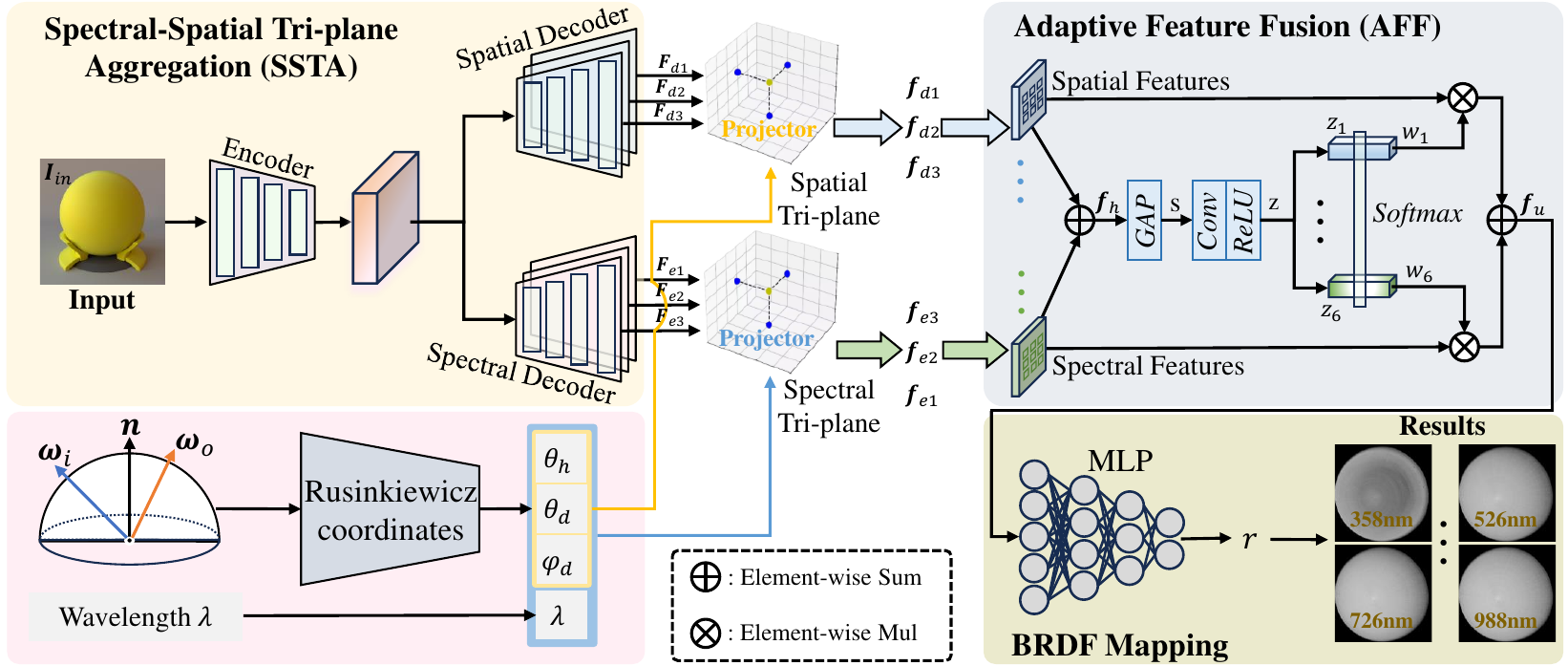}
	\end{overpic}
	\vspace{-0mm}
	\caption{Pipeline of our SpecGen.}
	\label{fig:main}
	\vspace{-0mm}
\end{figure*}

As shown in Fig.~\ref{fig:main}, given an input RGB image, the SSTA network leverages an encoder-decoder to extract two sets of triplanes: a spatial triplane that encodes the reflectance response of incident and outgoing light angles, and a spectral triplane that captures the reflectance responses jointly across angles and wavelengths. Next, we project wavelengths and incident-outgoing angles, represented in Rusinkiewicz coordinates~\cite{rusinkiewicz1998new}, onto the two triplanes to generate six feature vectors. The AFF module then fuses these features and produces a latent BRDF feature, which is used as input for our MLP-based BRDF mapping module to predict the corresponding spectral reflectance~$r$.

\subsection{Spectral-Spatial Tri-plane Aggregation}

\paragraph{Insight: K-Planes~\cite{fridovich2023k} for Dynamic Neural Radiance Field.} 
Dynamic neural radiance fields can be represented as a 4D function with respect to 3D spatial location $(x, y, z)$ and a 1D timestamp $t$. Therefore, K-Planes~\cite{fridovich2023k}, as a dynamic NeRF, models the dynamic scene radiance $c$ as
\begin{equation}
c = f\left(x, y, z, t\right).
\end{equation}
To decompose the static canonical scene and dynamic motion field, K-Planes~\cite{fridovich2023k} introduces a planar factorization that represents this 4D function using $C_{4}^{2}$ feature planes. Among these, three planes $\left\{ (x, y),\ (x, z),\ (y, z) \right\}$ encode spatial domain information, while the remaining three $\left\{ (x, t),\ (y, t),\ (z, t) \right\}$ capture spatial-temporal variations, thus enabling a clear decoupling of the static radiance field and temporal motion field.

\vspace{-3mm}
\paragraph{Spectral BRDF Representation} 
Based on Rusinkiewicz coordinates~\cite{rusinkiewicz1998new}, the spectral BRDF response can also be modeled as a 4D function of the incident-outgoing angles $(\theta_h, \theta_d, \varphi_d)$ and wavelength $\lambda$. Specifically, given incident and outgoing light directions $\boldsymbol{\omega_i}$ and $\boldsymbol{\omega_o}$, the half vector can be calculated as
$\boldsymbol{h} = \frac{\boldsymbol{\omega_i} + \boldsymbol{\omega_o}}{\|\boldsymbol{\omega_i} + \boldsymbol{\omega_o}\|}$.
The angle $\theta_h$ denotes the polar angle of $\boldsymbol{h}$ relative to the fixed surface normal $\boldsymbol{n} = (0, 0, 1)$; $\theta_d$ is the polar angle between the outgoing direction $\boldsymbol{\omega_o}$ and the half-vector $\boldsymbol{h}$; $\varphi_d$ is the azimuth angle around $\boldsymbol{h}$. Similar to the 4D function in Dynamic NeRF, the spectral BRDF response $r$ can be denoted as:
\begin{equation}
r = f\left(\theta_h,\theta_d,\varphi_d,\lambda\right).
\end{equation}

As shown in Fig.~\ref{fig:main}, we follow the decomposition in K-Planes~\cite{fridovich2023k} to decompose the 4D spectral response into two triplanes. The three planes $\{ (\theta_h, \theta_d), (\theta_h, \varphi_d), (\theta_d, \varphi_d) \}$, which depend solely on angular parameters, are designated as the spatial triplane. We denote these three planes as $\boldsymbol{F}_{d1}$, $\boldsymbol{F}_{d2}$, and $\boldsymbol{F}_{d3}$, respectively. Meanwhile, the remaining three planes $\{ (\theta_h, \lambda), (\theta_d, \lambda), (\varphi_d, \lambda) \}$ incorporate wavelength dependence and are referred to as the spectral triplane, denoted by $\boldsymbol{F}_{e1}$, $\boldsymbol{F}_{e2}$, and $\boldsymbol{F}_{e3}$. Since the spatial triplane is only related to incident-outgoing angles, it can be shared by both RGB and spectral BRDFs. As discussed in our RGB-spectral joint training, this decomposed representation enables the use of RGB data to enhance spectral BRDF generation.
\vspace{-3mm}

\paragraph{Image-to-Triplane Mapping}
We employ a CNN-based encoder-decoder network architecture to generate the spatial and spectral triplanes. As shown in Fig.~\ref{fig:main}, the network input is an RGB image of the target material rendered on a sphere. The output is a 6-channel feature map, which compose the two triplanes: spatial triplane $\left\{ \boldsymbol{F}_{d1}, \boldsymbol{F}_{d2}, \boldsymbol{F}_{d3} \right\}$ and spectral triplane $\left\{ \boldsymbol{F}_{e1}, \boldsymbol{F}_{e2}, \boldsymbol{F}_{e3} \right\}$.
To obtain the spectral BRDF value, we first calculate the coordinate $(\theta_h,\theta_d,\varphi_d,\lambda)$ and project this coordinate onto the two triplanes, leading to $6$ feature vectors with the dimension of $C$, \ie,

\begin{equation}
{\boldsymbol{f}_{{d1}}},{\boldsymbol{f}_{{d2}}},{\boldsymbol{f}_{{d3}}} = \left( {{\mathcal{P}}_{{\boldsymbol{F}_{d1}}}^{({\theta_h},{\theta_d})},{\mathcal{P}}_{{\boldsymbol{F}_{d2}}}^{({\theta_h},{\varphi_d})},{\mathcal{P}}_{{\boldsymbol{F}_{d3}}}^{({\theta _d},{\varphi _d})}} \right),
\end{equation}
\begin{equation}
{\boldsymbol{f}_{{e1}}},{\boldsymbol{f}_{{e2}}},{\boldsymbol{f}_{{e3}}} = \left( {{\mathcal{P}}_{{\boldsymbol{F}_{e1}}}^{({\theta _h},{\lambda})},{\mathcal{P}}_{{\boldsymbol{F}_{e2}}}^{({\theta _d},{\lambda})},{\mathcal{P}}_{{\boldsymbol{F}_{e3}}}^{({\varphi _d},{\lambda})}} \right),
\label{eq:project}
\end{equation}
where $\mathcal{P}$ denotes the projector operator. 
${\boldsymbol{f}_{{di}}}$ and ${\boldsymbol{f}_{{ei}}}$ represent $C$-dimensional feature vectors associated with the spatial and spectral parameters, respectively. 
The following section describes how these spatial and spectral features are fused to predict the BRDF response.

\subsection{Adaptive Feature Fusion (AFF)}
Unlike previous methods~\cite{fridovich2023k,li2025dual} that conduct feature fusion via dot product, we propose an Adaptive Feature Fusion (AFF) module, inspired by SKNet~\cite{li2019selective}, to dynamically select appropriate weights for fusing different features. As shown in Fig.~\ref{fig:main}, we first reshape the $C$-dimensional 1D feature vectors into 3D feature maps of size $1\times H \times W$. Then, we apply $3 \times 3$ convolutions and element-wise addition to promote feature interaction, resulting in fused hybrid features, denoted as~$\boldsymbol{f}_h \in \mathbb{R}^{1\times H \times W}$:
\begin{equation}
{{\boldsymbol{f}}_h} = \sum\limits_{i = 1}^3 {C{\rm{onv}}\left( {{{\boldsymbol{f}}_{ei}}} \right) + \sum\limits_{i = 1}^3 {C{\rm{onv}}\left( {{{\boldsymbol{f}}_{di}}} \right)} }.
\end{equation}
Next, we perform global average pooling ${\mathcal{GAP}}$ to obtain the channel statistics $s$, which capture the global context:
\begin{equation}
s = {\mathrm{GAP}}\left( {{\boldsymbol{f}_h}} \right) = \frac{1}{{H \times W}}\sum\limits_{i = 1}^H {\sum\limits_{j = 1}^W {\boldsymbol{f}_h(i,j)}}.
\end{equation}
To enrich feature representations, inspired by bottleneck structures~\cite{hu2018squeeze,li2024efficient}, we first expand $s$ into a descriptor $\boldsymbol{z} \in \mathbb{R}^{d}$ ($d=32$) using a $1 \times 1$ convolution followed by a ReLU activation. We then apply six parallel, branch-specific convolutions, each with kernel $\boldsymbol{\beta}_i$, to reduce the channel dimensionality, enhance inter-channel dependencies, and extract multi-branch feature representations. The process for the $i$-th branch is as follows:
\begin{equation}
z = \text{ReLU}\left( \text{Conv}\left( s \right) \right),
\end{equation}
\begin{equation}
z_i = \text{Conv}(z, \boldsymbol{\beta}_i) \quad \text{for} \quad i = 1, ..., 6.
\end{equation}
To adaptively weight different feature branches, we compute an attention score for each branch using a softmax-based mechanism. Specifically, the weight $w_i$ for the $i$-th branch feature $z_i \in \mathbb{R}$ is obtained as follows:
\begin{equation}
{w_i} = \text{Softmax}\left( \text{Reshape}\left( z_i \right) \right).
\end{equation}
Input spatial and spectral features are re-weighted by their corresponding adaptive scores and then aggregated to form the final fused output $\boldsymbol{f}_u \in \mathbb{R}^C$, which effectively captures the relative importance of different features during fusion:
\begin{equation}
{\boldsymbol{f}_u} = \text{Reshape}\left(\sum\limits_{i = 1}^3 \left({{{w}_i}{{\boldsymbol{f}}_{{d_i}}}}  +  {{{w}_{i+3}}{{\boldsymbol{f}}_{{e_i}}}}\right)\right) .
\end{equation}
where the reshape operator is used to transform the 2D feature map into a 1D feature vector. With this design, our AFF module effectively enhances feature interaction by adaptively selecting receptive fields for different features and assigning dynamic weights, thus enabling more accurate spectral BRDF reconstruction.

\paragraph{BRDF Mapping Module}
Given the fused feature ${\boldsymbol{f}_u} \in \mathbb{R}^C$, we use a three-layer MLP with ReLU activations to decode ${\boldsymbol{f}_u}$ and generate the spectral BRDF value $r$ at the coordinate $(\theta_h, \theta_d, \varphi_d, \lambda)$:
\begin{equation}
r = {\rm MLP}({\boldsymbol{f}_u}; \boldsymbol{\delta}),
\end{equation}
where $\boldsymbol{\delta}$ are the MLP parameters. With SSTA feature representation, AFF-based fusion, and this MLP mapping, we predict the spectral reflectance $r$ at each coordinate. Repeating this process over all directions and wavelengths, we reconstruct the full spectral BRDF, enabling flexible spectral image rendering under diverse illumination and geometry.

\subsection{RGB-Spectral Joint Training Strategy} 
The spatial and spectral decomposition introduced by the SSTA module is designed to address the scarcity of spectral BRDF data in generation tasks by leveraging the abundance of RGB BRDF data. Building on the SSTA module, we propose a joint RGB-spectral training strategy that utilizes both RGB and spectral BRDF data. Below, we detail the workflow and loss functions employed for training on both RGB and spectral BRDF data.

\begin{figure*}[ht]
	\begin{overpic}[width=0.99\linewidth]{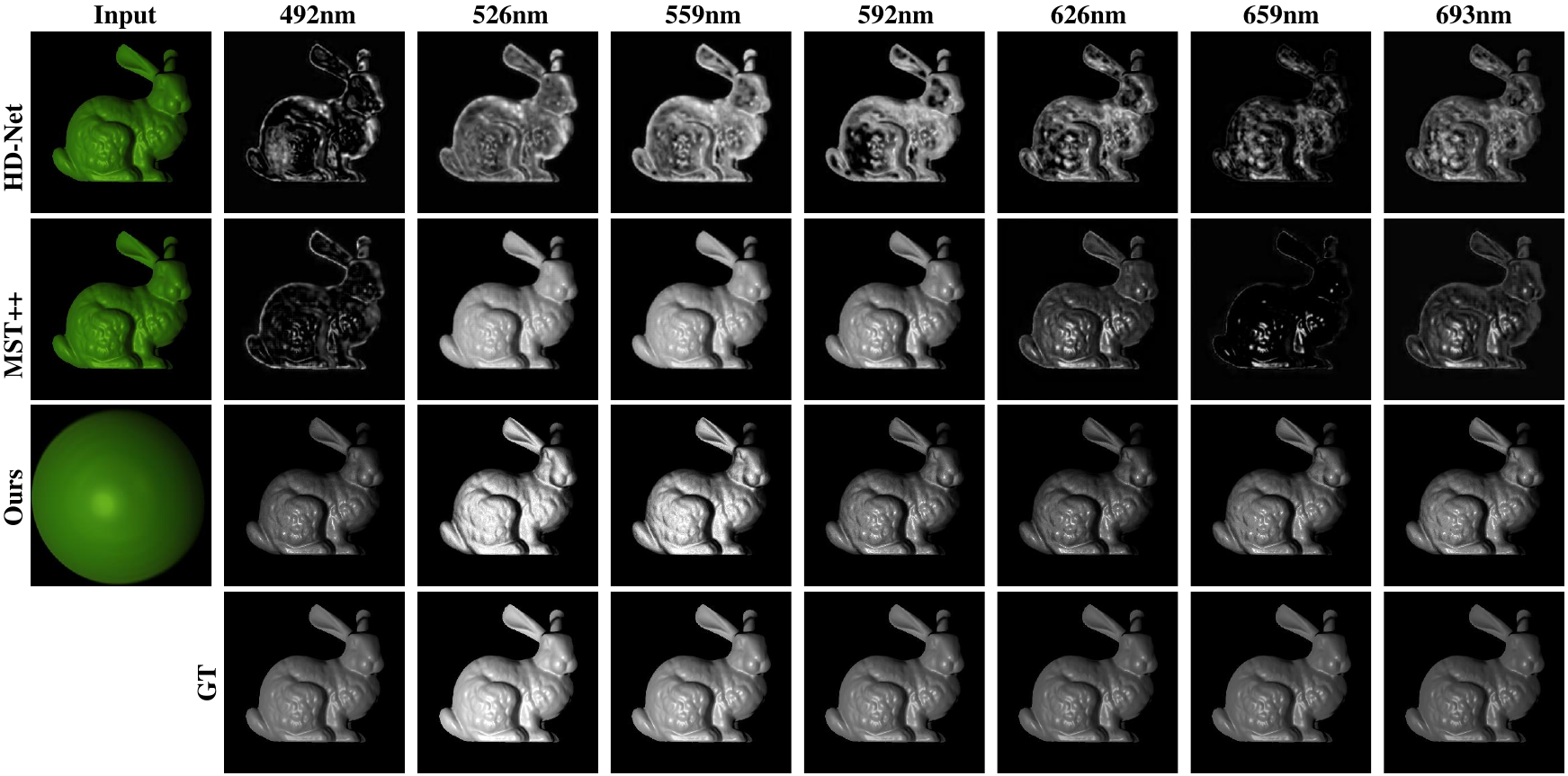}
		\put(-0.0,43.3){\color{black}{\fontsize{7.8pt}{1pt}\selectfont {\rotatebox{90}{\cite{cai2022mask}}}}}
        \put(-0.0,31.2){\color{black}{\fontsize{7.8pt}{1pt}\selectfont {\rotatebox{90}{\cite{hu2022hdnet}}}}}
	\end{overpic}
	\vspace{-1mm}
	\caption{Qualitative comparisons of spectral rendering, where our method takes the RGB sphere input image, generates the spectral BRDF, and renders the spectral bunny image. HSI baselines take the RGB bunny image and directly output the corresponding spectral image.}
	\label{fig: visual2}
	\vspace{-1mm}
\end{figure*}

\paragraph{Training on Spectral BRDF} 
Measured BRDFs often exhibit a high dynamic range (HDR), resulting in large numerical fluctuations, particularly in the specular components. To address this issue, we apply the $\mu$-law, a logarithmic-relative mapping, to compress the HDR data into a range that is more suitable for training:
\begin{equation}
    r' = \frac{\log(1+\mu\left| r \right|)}{\log(1+\mu)},
\end{equation}
where $r'$ denotes the scaled BRDF values, and $\mu$ is a compression parameter controlling the degree of compression. In our experiments, we set $\mu=255$ to achieve effective compression without significantly degrading data quality.

Given $N$ sampled coordinates $(\theta_h,\theta_d,\varphi_d,\lambda)$, we measured the difference between GT and estimated spectral response ${r}_i'$ and $\hat{r}_i'$ after $\mu$-law, \ie,
\begin{equation}
\mathcal{L}_{spec}=\frac{1}{N} \sum_{i=1}^{N} (\hat{r}_i'-r_i')^2.
\end{equation}
Meanwhile, we apply a total variation~(TV) loss $\mathcal{L}_{TV}$ to constrain the smoothness of generated spectral BRDF. The total loss on spectral BRDF can be represented as 
\begin{equation}
     \mathcal{L}_{s} = \mathcal{L}_{spec} + \alpha \mathcal{L}_{TV},
\end{equation}
where coefficient $\alpha$ is set to 2 for balancing the TV loss. 

\paragraph{Training on RGB BRDF} 
RGB BRDFs and spectral BRDFs share the same spatial tri-plane features ($\boldsymbol{f}_{di}$), but differ in their spectral tri-plane features ($\boldsymbol{f}_{ei}$). Specifically, for each coordinate $(\theta_h, \theta_d, \varphi_d, \lambda)$ in the spectral BRDF, we sample the corresponding spectral tri-plane feature using the projection operation described in Eq.~\eqref{eq:project}.
For the RGB BRDF case, the spectral feature is obtained by averaging all features along the $\lambda$ dimension of the spectral triplane at the location $(\theta_h, \theta_d, \varphi_d)$, \ie,
\begin{equation}
    \boldsymbol{\hat{f}}_{ei} = \frac{1}{M}\sum_{\lambda} \mathcal{P}_{{\boldsymbol{F}_{ei}}}^{(\theta_h, \theta_d, \varphi_d, \lambda)},
\end{equation}
where $M$ is the number of sampled wavelengths in the spectral triplane, and $\boldsymbol{\hat{f}}_{ei}$ denotes the spectral features for the RGB BRDF data. This averaging operation effectively converts the spectral response into a grayscale average reflectance. Given the spatial and spectral features, we predict the response $\hat{r}'$ and supervise it using the grayscale value of the RGB BRDF:
\begin{equation}
L_{\mathrm{RGB}} =
\frac{1}{N} \sum_{i=1}^N
\left( \hat{r}'_i - g'_i \right)^2
\end{equation}
where $g'$ is obtained by averaging the RGB BRDF values across channels and applying the $\mu$-law transformation.

To train SpecGen, we employ the loss functions $\mathcal{L}_{RGB}$ and $\mathcal{L}_{spec}$ for RGB and spectral BRDF data, respectively. This joint training strategy allows us to leverage large-scale RGB BRDF data, enhancing the model's generalization.

\vspace{-2mm}

\section{Experiment}
\subsection{Dataset and Evaluation Metrics}

We evaluate our method on two public BRDF datasets: the RGL dataset~\cite{dupuy2018adaptive} and the MERL dataset~\cite{matusik2003data}. RGL contains 51 real-world isotropic materials with both RGB and spectral BRDF measurements. MERL provides BRDFs for 100 real-world materials, parameterized in the Rusinkiewicz domain $(\theta_h,\theta_d,\varphi_d)$~\cite{rusinkiewicz1998new} using the $\mathbf{h}$ and $\mathbf{d}$ vectors. Each wavelength channel has a resolution of $90 \times 90 \times 180$, yielding 1,458,000 measurements across 195 wavelengths.
To enable meaningful perceptual comparison, we render predicted spectral BRDFs on a reference sphere under diverse illuminations and compute PSNR and SSIM~\cite{wang2002universal} against renderings generated from ground-truth BRDFs.


\subsection{Implementation Details}

We implement all experiments using PyTorch with the Adam optimizer~\cite{kingma2014adam}, setting $\beta_1=0.9$ and $\beta_2=0.999$. The learning rate is initialized at $1 \times 10^{-4}$, batch size is 65,536 (sampled points from multiple materials), and the learning rate decays by half every four epochs for stable convergence. Our model trains for 20 epochs on an RTX 4090 GPU (approx. 20 hours). We use 36 materials from the RGL isotropic dataset~\cite{dupuy2018adaptive} for training and 12 for testing. For both, we sample 5,120,000 points per material (about 1/10 of total) and apply $\mu$-law compression to reduce dynamic range. Spectral and spatial triplane dimensions are initialized as [90, 90, 180, 39], with 39 for wavelength channel.

\subsection{Evaluation on Spectral BRDF Generation}
\paragraph{Baselines} 


Our method reconstructs spectral BRDFs of diverse materials from a single input image, enabling detailed spectral rendering across wavelengths. In contrast, hyperspectral image reconstruction methods aim to recover spectral images directly, with a different modeling focus. Despite this distinction, both generate spectral representations that can be evaluated for fidelity. Therefore, we assess the quality of rendered spectral images using PSNR and SSIM~\cite{wang2002universal} under controlled conditions. We further compare our approach with state-of-the-art hyperspectral methods, including HD-Net~\cite{hu2022hdnet} for high-resolution recovery and MST++~\cite{cai2022mask}, a mask-guided transformer for spectral reconstruction.

\begin{figure*}[ht]
	\begin{overpic}[width=0.99\linewidth]{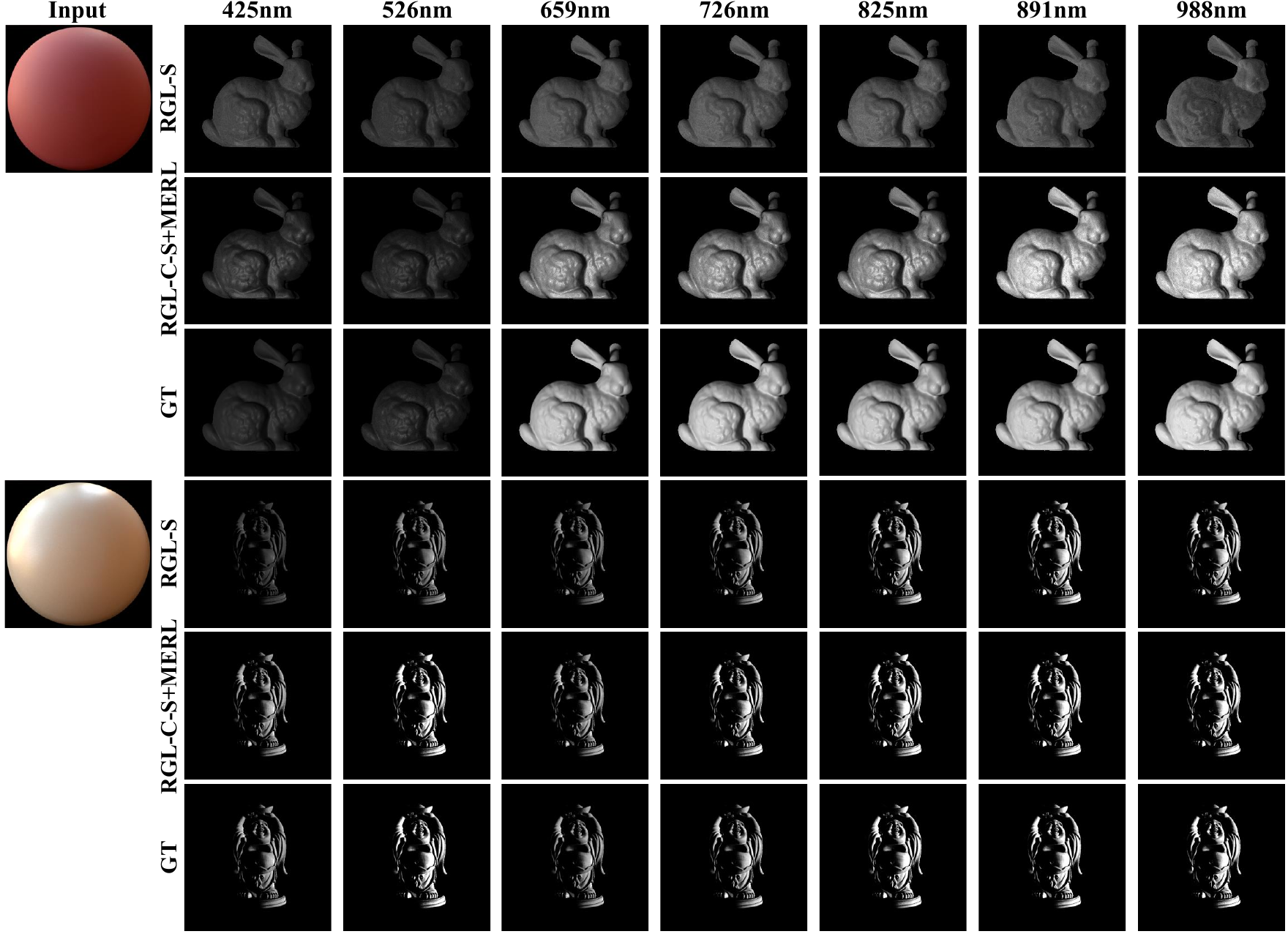}
	\end{overpic}
	\vspace{-1mm}
	\caption{Qualitative comparisons of rendering results using only spectral BRDF dataset RGL-S vs. spectral plus RGB BRDF dataset RGL-C-S+MERL, where top and bottom parts denote two different test materials. With the help of our training strategy building upon our SSTA network, we can leverage RGB BRDF data to boost the spectral rendering quality.}
	\label{fig: visual1}
	\vspace{-1mm}
\end{figure*}
\vspace{-3mm}
\paragraph{Qualitative and Quantitative Assessment} 
As shown in Table~\ref{tab: hyperspectral}, for three representative materials we compare our method with baselines using sphere images rendered under both environmental and distant lights, generating spectral images at nine uniformly spaced wavelengths. Our method consistently outperforms in multispectral reconstruction across diverse lighting conditions, with an average PSNR gain of over 8\,dB. As shown in Fig.~\ref{fig: visual2}, it yields reconstructions significantly closer to the ground truth than HSI methods under both parallel and varied environmental lighting. Supplementary material provides additional qualitative results that further confirm its superior generalization ability and robustness in complex lighting scenarios, with accurate rendering across the visible and near-infrared spectra.



\vspace{-0mm}
Unlike existing HSI methods, typically limited to the 400–700 nm visible spectrum and constrained to fixed-shape outputs, our method supports a significantly broader spectral range covering both visible and near-infrared bands, and generates high-fidelity spectral BRDFs that enable realistic multispectral rendering of objects with diverse shapes and materials.
Supplementary material provides results for various materials at representative visible and near-infrared wavelengths, demonstrating strong generalization and fidelity. As shown in Fig.~\ref{fig: visual2}, using the spectral BRDF from a sphere image, our method can render a bunny model with superior reconstruction across a wide wavelength range.


\paragraph{Robustness Assessment} Materials in real-world scenes face diverse environmental lighting. A robust reconstruction should preserve fidelity under such variations. As shown in Table~\ref{tab: environmental_light}, we evaluate our model under three lighting setups: Outdoor Env. map, Indoor Env. map1, and Indoor Env. map2. Only the last was seen in training. We reconstruct images at 20 wavelengths and report the average PSNR and SSIM to assess quality. Strong performance on unseen lighting shows the model's robustness. Supplementary material provides qualitative results showing realistic, accurate renderings across wavelengths that closely match the GT.

\begin{table}
    \centering
    \caption{Quantitative comparisons on spectral rendering between our method and existing hyperspectral image reconstruction methods under environmental lights (\textbf{top}) and distant lights 
    (\textbf{middle}).}
    
    \resizebox{0.48\textwidth}{!}{{\begin{tabular}{l@{\hskip 0.1in}c@{\hskip 0.1in}c@{\hskip 0.1in}c@{\hskip 0.1in}c@{\hskip 0.1in}c@{\hskip 0.1in}c@{\hskip 0.1in}}\toprule 
    
\multirow{2}{*}{Material}  & \multicolumn{2}{c}{HD-Net~\cite{hu2022hdnet}} & \multicolumn{2}{c}{MST++~\cite{cai2022mask}}& \multicolumn{2}{c}{\textbf{Ours}}\\
\cmidrule(lr){2-3}\cmidrule(lr){4-5}\cmidrule(lr){6-7}
   & PSNR\textuparrow  & SSIM\textuparrow&PSNR\textuparrow  & SSIM\textuparrow&PSNR\textuparrow  & SSIM\textuparrow\\
 \toprule
acrylic-felt-yellow &  18.64  & 0.6737& 20.85   & 0.7103& 29.39   & 0.9303\\
ilm-l3-37-dark-green &  31.32  & 0.7912&  28.27& 0.7448& 47.36 & 0.9677\\
cc-ibiza-sunset& 24.94 & 0.6690 & 26.63   & 0.7010& 38.19  & 0.8951\\
 \toprule
acrylic-felt-yellow & 25.01 & 0.5989& 29.12  & 0.7476& 29.47  & 0.8292\\
ilm-l3-37-dark-green & 33.03 & 0.9852 &  32.51& 0.9514& 39.57  &  0.9671\\
cc-ibiza-sunset& 25.53& 0.9627 & 24.77  & 0.9491& 27.34  & 0.9155\\

\toprule
Average & 26.41 & 0.7801& 27.03 & 0.8007&\textbf{35.22} &\textbf{0.9175}\\
\bottomrule

\end{tabular}\par}}
\label{tab: hyperspectral}
\vspace{-1mm}
\end{table}

 \begin{table}[t!]
    \centering
    \caption{Ablations on the robustness of our method under different environmental lights. \textcolor{gray}{Gray} cells indicate environmental lights present in training datasets. Our method shows generalization even in environmental light that is not covered by the training dataset.}
    \resizebox{0.48\textwidth}{!}{{\begin{tabular}{l@{\hskip 0.1in}c@{\hskip 0.1in}c@{\hskip 0.1in}c@{\hskip 0.1in}c@{\hskip 0.1in}}\toprule 
    
\multirow{2}{*}{Env. Light} & \multicolumn{2}{c}{acrylic-felt-yellow} & \multicolumn{2}{c}{ilm-l3-37-matte}\\
\cmidrule(lr){2-3}\cmidrule(lr){4-5}
   & PSNR\textuparrow  & SSIM\textuparrow&PSNR\textuparrow  & SSIM\textuparrow\\
 \toprule
Outdoor Env. map & 39.44 & 0.9609& 41.95  & 0.9847\\
Indoor Env. map1& 38.28 & 0.9551 & 41.06 & 0.9781\\
\textcolor{gray}{Indoor Env. map2} & 36.72 & 0.9705 & 40.39  & 0.9812\\

\toprule
Average & 38.15 & 0.9622& 41.13 & 0.9813 \\
\bottomrule
\label{tab: environmental_light}
\end{tabular}\par}}
\vspace{-5mm}
\end{table}

\subsection{Ablation study}
\paragraph{Training strategy}
To assess our RGB–spectral joint training strategy for mitigating data scarcity, we compare models trained with and without RGB BRDF data. We use two training sets: (1) RGL-S, containing 36 spectral BRDFs from RGL~\cite{dupuy2018adaptive}; and (2) RGL-C-S+MERL, which adds 51 RGB BRDFs from RGL and 100 from MERL~\cite{matusik2003data}. SpecGen is trained on each and evaluated on 12 unseen spectral BRDFs. As shown in Table~\ref{tab: RGL-C-S+MERL}, adding RGB data improves spectral reconstruction by nearly 0.9\,dB PSNR on average. Figure~\ref{fig: visual1} provides qualitative results for two representative materials at seven distinct wavelengths, showing that RGB–spectral joint training yields more accurate reflectance predictions and produces rendered images that are perceptually and quantitatively closer to the GT. These results highlight the effectiveness of leveraging abundant RGB BRDF data as complementary supervision to enhance the fidelity and consistency of spectral BRDF reconstruction across the spectrum.

 \begin{table}
    \centering
    \caption{Evaluation on the effectiveness of boosting spectral BRDF generation via RGB BRDF data on 12 test spectral BRDF data. We calculate PSNR and SSIM of spectral rendering of a sphere using GT and generated spectral BRDFs under distant light.}
    \resizebox{0.48\textwidth}{!}{
    {\begin{tabular}{l@{\hskip 0.1in}c@{\hskip 0.1in}c@{\hskip 0.1in}c@{\hskip 0.1in}c@{\hskip 0.1in}c}\toprule
    
\multirow{2}{*}{Material}& \multicolumn{2}{c}{RGL-S} & \multicolumn{2}{c}{RGL-C-S+MERL}\\
\cmidrule(lr){2-3}\cmidrule(lr){4-5}
   & PSNR\textuparrow  & SSIM\textuparrow & PSNR\textuparrow  & SSIM\textuparrow  \\

 \toprule

acrylic-felt-green& 33.28 & 0.8651 & 34.50 & 0.8793 \\
acrylic-felt-orange& 33.73 &0.9292 & 32.46 & 0.8770 \\
acrylic-felt-purple& 31.21& 0.7196 & 31.87 & 0.8135\\
acrylic-felt-yellow& 30.70& 0.9104& 28.08 & 	0.8638\\
leaf-maple& 34.67& 0.7621& 33.62 & 0.7484 \\
ilm-l3-37-matte& 29.70& 0.8740& 28.56 & 0.8881 \\
cc-ibiza-sunset& 28.98& 0.7976& 31.81 & 0.9354 \\
ilm-l3-37-metallic& 25.37& 0.5473& 	27.02 & 0.8147 \\
colodur-kalahari-2a& 28.43& 0.9157& 	29.07 & 0.8761 \\
chm-light-blue& 31.29& 0.8416&32.61 & 0.8490 \\
spectralon& 25.38& 0.8983& 27.80& 0.8420 \\
colodur-napoli-4f& 36.48& 0.8881& 42.35 &0.9622 \\
 \toprule
Average& 30.77&  0.8291& \textbf{31.65}& \textbf{0.8625} \\

\bottomrule
\label{tab: RGL-C-S+MERL}
    \end{tabular}\par}}
    \vspace{-4mm}
\end{table}

 \begin{table}
    \centering
    \caption{Ablations on triplane feature fusion modules, including Hadamard product used in K-Planes~\cite{fridovich2023k} and our AFF module.}
    \resizebox{0.48\textwidth}{!}{
    {\begin{tabular}{l@{\hskip 0.2in}c@{\hskip 0.2in}c@{\hskip 0.2in}c@{\hskip 0.2in}c@{\hskip 0.2in}}\toprule
     \multirow{2}{*}{Method}& \multicolumn{2}{c}{RGL-S}& \multicolumn{2}{c}{RGL-C-S+MERL} \\
\cmidrule(lr){2-3}\cmidrule(lr){4-5}\
    & PSNR\textuparrow  & SSIM\textuparrow& PSNR\textuparrow  & SSIM\textuparrow \\
 \toprule
Hadamard~\cite{fridovich2023k} & 27.00 & 0.7837& 26.83 & 0.8159\\
AFF (Ours) & \textbf{30.77}  & \textbf{0.8291} & \textbf{31.65} & \textbf{0.8625}\\
\bottomrule
\label{table: aff}
\end{tabular}\par}}
\vspace{-4mm}
\end{table}

\vspace{-3mm}
\paragraph{Adaptive Feature Fusion Module}
Previous studies~\cite{fridovich2023k} have highlighted the effectiveness of the Hadamard product in triplane feature fusion. To evaluate whether our proposed Adaptive Feature Fusion (AFF) module offers a more efficient alternative, we conduct a correlation ablation study in Table~\ref{table: aff}. The ablation results demonstrate a significant advantage of our AFF in multiplanar feature fusion, achieving an average PSNR improvement of 4.3\,dB across two test datasets compared to the Hadamard product strategy, showing the rationality and strength of the adaptive selection fusion mechanism in our AFF module.

\vspace{-1mm}
\section{Limitations and Future Work}
\vspace{-1mm}
Despite its strong empirical performance, SpecGen exhibits several limitations. First, the current framework assumes spherical input geometry. Relaxing this assumption to accommodate arbitrary shapes may lead to measurable accuracy degradation. Second, materials with pronounced specular behavior are highly sensitive to illumination variations, complicating reliable recovery of their spectral BRDFs. Third, naively extending RGB-based BRDF generation methods to the spectral domain substantially increases model complexity, undermining the fairness and interpretability of cross-method comparisons. Finally, while SpecGen reconstructs full BRDFs from sparse samples, the limited size of existing spectral BRDF datasets constrains its upper-bound performance. We plan to investigate the performance of our method once more spectral BRDFs are available, and compare with more baselines in our future work.
\vspace{-2mm}
\section{Conclusion}

We propose SpecGen, the first framework that generates spectral BRDFs from a single RGB sphere image, enabling photorealistic spectral rendering under arbitrary illuminations and geometries. To address the scarcity of spectral BRDF data, SpecGen leverages abundant RGB BRDFs through a SSTA that decouples reflectance modeling into spectral and spatial domains. Moreover, our AFF module integrates multi-plane features more effectively than fixed fusion operators. Extensive experiments on public spectral BRDF datasets and downstream spectral rendering tasks demonstrate the strong performance of our SpecGen.

\section*{Acknowledgements}
This work was supported by the National Natural Science Foundation of China (Grant No. 62472044, U24B20155),
Hebei Natural Science Foundation Project No. 242Q0101Z, Beijing-Tianjin-Hebei Basic Research Funding Program No. F2024502017, and the Youth Innovative Research Team of BUPT.

{
    \small
    \bibliographystyle{ieeenat_fullname}
    \bibliography{main}
}

\end{document}